\documentclass[Afour,sageh,times]{sagej}

\usepackage{moreverb}
\usepackage{url}
\usepackage[colorlinks,
            bookmarksopen,
            bookmarksnumbered,
            citecolor=red,
            urlcolor=red]{hyperref}
\usepackage[compact]{titlesec}
\usepackage{paralist}
\usepackage[normalem]{ulem}
\usepackage{tikz}
\usetikzlibrary{tikzmark}

\def\shownotes{0} 
\newcommand{\xxnote}[3]{}
\if\shownotes1
  \usepackage[dvipsnames]{xcolor}
  \renewcommand{\xxnote}[3]{\textcolor{#2}{#1: #3}}
\fi

\def\volumeyear{2018}

\begin{document}

\titlespacing{\subsection}{0pt}{0.5ex}{0.5ex}
\titlespacing{\subsubsection}{0pt}{0.5ex}{0.5ex}

\runninghead{Newman et al.}

\title{HARMONIC: A Multimodal Data Set of Assistive Human-Robot Collaboration}

\author{Benjamin A. Newman\affilnum{*}\affilnum{1}, Reuben M. Aronson\affilnum{*}\affilnum{1} \\Siddhartha S. Srinivasa\affilnum{2}, Kris Kitani\affilnum{1}, Henny Admoni\affilnum{1}}

\affiliation{\affilnum{*}Denotes equal contribution\\
\affilnum{1}Robotics Institute, Carnegie Mellon University, Pittsburgh, PA\\
\affilnum{2}University of Washington, Seattle, WA}

\corrauth{Benjamin A. Newman,
Robotics Institute,
Carnegie Mellon University,
Pittsburgh, Pennsylvania 15213}

\email{newmanba@cmu.edu}

\begin{abstract}
We present the Human And Robot Multimodal Observations of Natural Interactive Collaboration (HARMONIC) data set. This is a large multimodal data set of human interactions with a robotic arm in a shared autonomy setting designed to imitate assistive eating. The data set provides human, robot, and environmental data views of twenty-four different people engaged in an assistive eating task with a 6 degree-of-freedom (DOF) robot arm. From each participant, we recorded video of both eyes, egocentric video from a head-mounted camera, joystick commands, electromyography from the forearm used to operate the joystick, third person stereo video, and the joint positions of the 6 DOF robot arm. Also included are several features that come as a direct result of these recordings, such as eye gaze projected onto the egocentric video, body pose, hand pose, and facial keypoints. These data streams were collected specifically because they have been shown to be closely related to human mental states and intention. This data set could be of interest to researchers studying intention prediction, human mental state modeling, and shared autonomy. Data streams are provided in a variety of formats such as video and human-readable \texttt{CSV} and \texttt{YAML} files. 
\end{abstract}


\maketitle

\section{Introduction}

In human-robot collaborations, robots need to perceive, understand, and predict the effects of their own actions as well as the actions of their human partners. This is especially important for assistive robots, which perform actions toward a (sometimes implicit) human goal. To successfully produce these assistive actions, the robot system must perceive, understand, and predict human mental states (the human's goals, intentions, and future actions, often unknown to external observers) that determine what assistance the robot should provide. 

Concretely, when people complete physical tasks, their external behaviors---such as their eye gaze---can reveal insights about their internal mental states. An assistance system that can understand how these behaviors relate to the task can predict what objects and locations of a visual scene the human deems to be task relevant. The system can also use these behaviors to determine whether or not interactions with these objects or locations will take place, and qualities that describe these interactions. This information is not known to the system prior to completing a task, and is not relayed to the system by the human via traditional means (e.g. verbal or written communication). Thus understanding these mental states in order to assist the human requires perceiving and interpreting the human's behavior during human-robot collaborations.

\begin{figure}
    \centering
    \includegraphics[width=0.65\columnwidth]{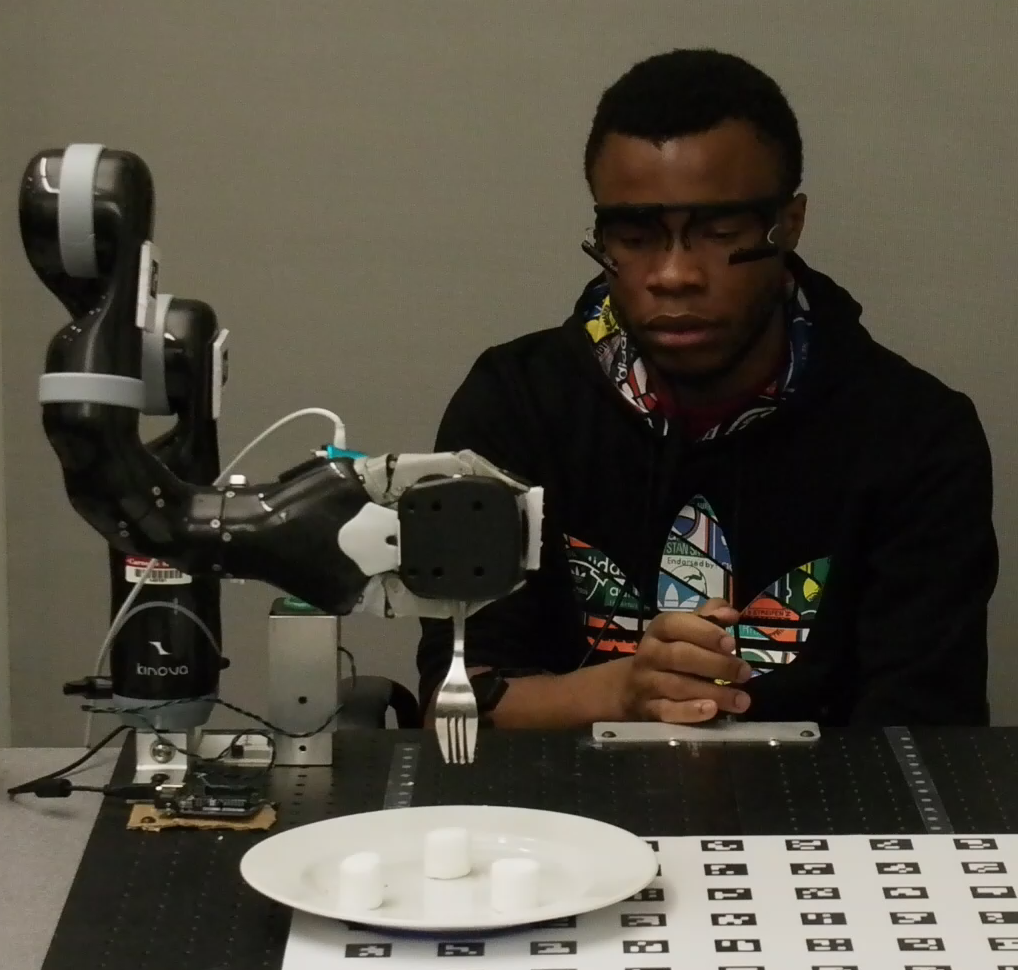}
    \caption{The HARMONIC data set provides multimodal human, robot, and environmental data collected during an assistive human-robot collaboration.}
    \label{fig:task}
\end{figure}

One example of a behavior that has been well studied in physical tasks is eye gaze. People almost exclusively fixate their eye gaze on objects or locations involved in their current task \citep{hayhoe05}, thereby ignoring task irrelevant parts of a scene. Should these objects or locations require a direct interaction, people fixate their gaze on these objects and locations prior to moving their hands to complete the interaction \citep{land01}, thus revealing the intended interaction object in advance of any physical contact. Gaze  also  lingers on key points in the task, such as obstacles, revealing certain landmarks of manipulation \citep{Johansson2001}. Additionally, people gaze at objects before uttering verbal references, which others can use to disambiguate and predict speech \citep{admoni14cogsci,boucher12}.

Other human behaviors can also reveal current mental states. Electromyography (EMG) signals, which record the electrical stimulation of muscle fibers, can indicate what action people are attempting to complete with their hands. 

Additionally, pupil size has been correlated with cognitive load \citep{beatty82,Krejtz2018,Bednarik2018}, and understanding current human body posture can both reveal desired tasks and help to avoid potentially dangerous collisions \citep{Mainprice2015}.

In this paper, we present the Human And Robot Multimodal Observations of Natural Interactive Collaboration (HARMONIC) data set. The HARMONIC data set contains human, robot, and environment data collected during the human-robot collaborative task (Figure \ref{fig:task}). In this task, people control an assistive robot arm to pick up bites of food in a simple eating scenario. The 6 degree of freedom Kinova Mico robot arm is controlled in three dimensions via  a 2 axis joystick and manual mode-switching. In some cases the robot provides additional assistance through shared autonomy \citep{javdani17ijrr}. 

Though the data were collected during an assistive eating task, their usefulness extends beyond the specific domain of eating. The manual condition can be used to study human teleoperation in the general case, for example with tasks using simplified grippers such as vacuum tooling. When combined with the shared autonomy conditions, these data can be used to study co-manipulation across individuals and varying levels of robot agency. Included in the data are a wide array of nonverbal behaviors situated in a real-world task defined with a clear goal and thus, is relevant for a variety of human-robot collaborations. 

Human behavioral data include egocentric RGB videos, eye gaze positions relative to these videos, infrared (IR) videos of both eyes, stereo, third-person video of the participant, and EMG recordings on the joystick-controlling arm. Robot related data include joystick control inputs from the user, the control input and belief distribution calculated by the assistance algorithm, and the robot position. Environmental data include the 3D locations of the food morsels as well as the locations of fiducial markers. Further information as well as an explanation as to how to access these data is offered in the following sections.


Our data set will help researchers study the complex human-robot dynamics of assistive teleoperation, which can vary across individual and across different levels of robot autonomy. For example, researchers could use this data set to learn correlations between eye gaze and joystick control, in order to improve the goal inference predictions made by shared autonomy algorithms. Others might be interested in modelling and forecasting the dynamics of joystick inputs under differing amounts of robot assistance. Previous research using similar data has proposed identifying unexpected events (\textit{e.g.}, human errors or task failure) by learning a normative gaze behavior model and identifying anomalies \citep{aronson18rss_fja}; the higher quality data provided in this dataset could continue this line of research as well as extend it to situations where the robot provides variable levels of assistance within a unified framework. 

\section{Prior Work}

\subsection{Human Interaction for Robotic Control}

Eye gaze, EMG, and body pose have all been useful signals for robotic control. Since eye gaze is a rich signifier of intention during manipulation, both by hand \citep{hayhoe05, Land2001, Johansson2001} and by robot \citep{aronson18}, its use has been explored through numerous robotic collaboration settings, including anticipating which object a user will request \citep{huang16}, and triggering assistive aid during autonomous driving \citep{Braunagel15}. Electromyography signals have been used for robot control \citep{Artemiadis10} and task monitoring \citep{DelPreto18}. 

Additionally, there has been work in learning and leveraging human policies (using keyboard input) \citep{reddyRSS18, reddyARXIV18} and attention models (using keyboard input and eye gaze) \citep{zhangECCV18} for both assisted and shared robot control in Atari games in an arcade learning environment \citep{atari}. HARMONIC provides a more realistic environment for studying such interactions. By making this data set available, we intend to enable further research into these control methods.

\subsection{Multimodal Data Sets in Human Robot Interaction}

Multimodal data sets have garnered interest in many different communities, such as psychology \citep{xu13}, computer vision \citep{adl, char, char-ego, epic, gtea, shu16}, human-robot interaction \citep{azgara16, benyoussef17, jayagopi13, sheikhi12, stefanov16}, and natural language processing \citep{bastianelli14}. These data sets, though, can be difficult to collect at a large scale. This can be due to the increasing engineering demand required with each additionally desired modality, physically collocating robots and humans, and the need to respect humans' privacy rights. This leads to many multimodal data sets including either few participants or few data modalities. In addition, these data sets are rarely designed to study direct, physical human robot collaborations in which the human and robot act in similar spaces. HARMONIC gives researchers the opportunity to study direct human robot collaboration in the form of a large scale data set in both the number of available modalities as well as the number of participants. Here, we compare how HARMONIC relates to other multimodal human robot interaction in order to illustrate these distinctions and the potential use of HARMONIC. 

\subsubsection{Robots in Conversational Settings}
The majority of publicly released HRI data sets study the inclusion of robots as conversational partners. To successfully incorporate robots as part of a social conversation, it is necessary to perceive  human behavior, understand how this relates to the conversation, and be able to synthesize similar behavior in order to keep the conversation flowing smoothly. Much of this work surrounds determining the human's visual focus of attention (VFOA) \citep{jayagopi13, sheikhi12}. In these works, VFOA is a discrete representation of eye gaze estimated from the user's head position. Other data sets are designed to capture unscripted conversations with a robot \citep{benyoussef17} by capturing conversations through a robot's third person video recorder. In all of these works, no signal specific sensors (e.g. an eye gaze camera) were used in order to capture specific human behaviors (e.g. eye gaze). 

Other conversational data sets have a linguistic focus \citep{bastianelli14}. This work designs an interaction in which a human commands a robot to perform a specific task, and contains many different views of the language spoken. Due to the focus on verbal communication, this data set does not give researchers the ability to understand how nonverbal behaviors may be utilized in order to understand the intent behind the human's command. 

Finally, perhaps the most similar data set to HARMONIC (in terms of data streams collected) again focuses on predicting the VFOA during a conversation \citep{stefanov16}. Unlike other works focusing on VFOA, this data set does capture eye gaze explicitly through the use of a Tobii eye tracker \citep{tobiiwebsite}. Thus, this work studies how gaze changes between structured and unstructured conversation with and without the presence of a robot. This data set is not designed, however, to study these nonverbal behaviors in physical collaborations. 

In all of these situations, the behaviors collected for analysis are centered around non-collaborative tasks. HARMONIC provides the opportunity to study how these behaviors may be interpreted in order to provide better assistance during collaborative tasks.

\subsubsection{Robot as Student}
The Multimodal Human-Robot Interaction Dataset  \citep{azgara16} is  designed for interactive object learning through human guidance. This data set presents a situation in which a human uses a small number of task specific behaviors in order to teach the robot about object models. This data set is intended to instruct the robot by leveraging a human's innate teaching ability, as opposed to studying physical human-robot collaboration. 

Humans teaching robots has also been studied in psychology \citep{xu13}. Here, researchers studied how humans' eye gaze patterns changed as a robot displayed gaze patterns that were designed to emulate the gaze patterns displayed by people who employ different styles of learning. Again, this task is different from a direct collaboration such as our shared autonomy task. Additionally, this data set does not seem to be publicly available. 

\subsection{Machine Learning and Computer Vision}
Surprisingly, our data set is similar to those from the machine learning and computer vision communities. The tasks studied in these data sets often include non-scripted, egocentric videos of daily activities \citep{epic}, gaze prediction for egocentric videos \citep{gtea}, action recognition in third person video \citep{adl, char}, relating first and third person videos as a proxy for theory of mind \citep{char-ego}, and learning about human social affordance from a third person view \citep{shu16}. These data sets include large amounts of potentially relevant data for human robot collaboration, but most importantly, they do not contain interactions with a robot. While these data sets may be useful for an initial understanding of human behavior, they do not provide insights into how these behaviors manifest in human robot collaborations. 

\section{Data Collection Procedure}

This section presents a brief overview of the user study and robot system in order to explain the conditions under which the data streams were recorded. 

\subsection{Participants}
Twenty-four participants (13 female) were recruited from the Pittsburgh area. Seventeen were between the ages of 18--24, four between 25--30, one between 31--35, and two between 41--45.
The participant pool was screened for prior experience using this robot arm in similar studies and, thus, were novices at the task. The experiment took place in the Human And Robot Partners (HARP) Lab on the Carnegie Mellon University campus. Participants were compensated \$15 for one and a half hours of their time. 

\subsection{Protocol}

Participants controlled a robot arm, attempting to position a fork above one of three marshmallows placed on a plate (see Fig.~\ref{fig:task}). They controlled a robot with a two-axis joystick using modal control: the joystick's two axes moved the end-effector of the robot in $x$ and $y$, $z$ and yaw, or pitch and roll. A joystick button allowed participants to cycle between control configurations when pressed for less 500 milliseconds. When the task was completed (that is, once a participant was satisfied with the fork's position or had given up on the task), the participant held down the same joystick button for longer than 500 milliseconds. This action triggered an autonomously executed plan in which the robot moved down to the height of the plate and speared the marshmallow (conditional on the proper positioning of the fork). Finally, the robot arm moved into a serving position near the participant's mouth. This concluded the trial, and the robot automatically reset to the starting configuration. 

Participants were given a brief introduction as to the purpose of the study and then began a five-minute familiarization period, in which they controlled the robot in teleoperation mode and data were not recorded. Next, participants were fitted with eye gaze and EMG sensors (described below). They performed the task five times in sequence for each of four assistance modes (described in the next section). Assistance mode order was fully counterbalanced among participants. After each block of five trials, participants were given a brief survey to record their subjective perceptions about the algorithm. Once the final survey was completed, participants were presented with a survey that compared all conditions through ranked preference as well as free response.

\subsection{Assistance Conditions}
Participants operated the robot in each of four different assistance conditions: fully teleoperated, two different levels of assistance according to the shared autonomy framework  \citep{javdani17ijrr}, and a fully autonomous robot.

The following is a brief description of how assistance is calculated; a full description is available in a prior publication  \citep{javdani17ijrr}. The combined human-robot system is modeled as a Partially Observable Markov Decision Process (POMDP)  \citep{KAELBLING199899,sondick}, where the participant's goal is represented as one unknown member of a small set of possible goals. Participant inputs via joystick are treated as observations. The algorithm assumes that the user is noisily optimizing a cost function parameterized by their unknown goal. Therefore, the Maximum Entropy Inverse Optimal Control (MaxEntIOC) \citep{ziebart2008maximum} framework can be used to evaluate a belief distribution over the known goal set. From this belief state, the overall POMDP is solved by applying the QMDP  \citep{LITTMAN1995362} approximation, which has proved reliable for similar shared control scenarios. Our implementation changed the original formulation slightly in order to remove the inherent living reward, which can cause the robot to converge on a goal even in the absence of any positive joystick actuation. The resulting robot action consists of a computed assistive action based on the inferred user goal distribution combined with the original applied user action.

To provide different assistance levels, the shared autonomy transition function was modified slightly from prior work. In Javdani et al.\ \citep{javdani17ijrr}, the given transition function applies both user and robot control as determined by $a_{applied} = u + a$.

In order to adapt the amount of user control, the applied action was parameterized by a value $\gamma$: $a_{applied} = ( 1 - \gamma ) u + \gamma a$, which trades off between the relative strengths of the user command and the robot assistance. Note that the original shared autonomy procedure would correspond to the case $\gamma = 0.5$ and normalizing the vector $a_{applied}$.

The four conditions corresponded to four different levels of $\gamma$:
\paragraph{Direct teleoperation, $\gamma = 0$.} The assistance signal $a$ was computed but completely discarded, so the user had full manual control over the robot.
\paragraph{Low assistance, $\gamma = 0.33$.} The assistance signal was combined with the direct user control, with the user signal weighted double. 
\paragraph{High assistance, $\gamma = 0.67$.} The assistance signal was combined with direct user control, but the assistance signal was more highly weighted.
\paragraph{Autonomous robot control, $\gamma = 1$.} The user control signal was not passed through to the robot control. It was used for goal inference only, and the robot was autonomously controlled based on its goal inference results.

\subsection{Sensors}

\subsubsection{Eye gaze}
Participant eye gaze direction was captured by a Pupil Labs Pupil \citep{pupillabswebsite, PupilLabsAcademic} sensor. This sensor consists of a glasses-like frame with two infrared cameras with infrared illumination mounted below each eye for dark pupil tracking, plus a third RGB camera oriented outward to capture egocentric video. The eye cameras capture video at 120 Hz, and pupil labs software detects the pupil pixel center. Before data were captured, the pupil locations and world camera videos were calibrated by asking the participant to look at the center of the marker held in front of them by the researcher (``manual marker calibration''). This calibration routine was recorded for most participants and is made available in the \texttt{calib} folder. The calibration is verified between each condition by asking participants to look at particular places in the scene. These checks are recorded and made available in the \texttt{check} folders.

\subsubsection{EMG}
Participant muscle activation while controlling the joystick was captured using a Myo sensor \citep{myowebsite}. Due to initialization failures, these data are only available for about 20\% of the runs (see Table~\ref{tab:coverage} for full details). It consists of the EMG message, denoting the activation of eight individual EMG sensors, the ORI message, denoting the orientation of the arm in roll/pitch/yaw, and the IMU message, denoting the readings of the IMU attached to the armband.

\subsubsection{External video}
Participant behavior was captured using a Stereolabs \citep{stereolabswebsite} ZED camera. Left and right videos are stored as separate \texttt{MP4} files. The ZED camera was placed on a tripod at approximately the same (marked) location for each trial in order to capture a full-on view of the participant and occasional views of the scene. ZED videos are available for the 10 participants who consented to their images being released. In all cases, offline skeleton and face tracking information is available.

\section{Descriptive Statistics}
This data set consists of 480 trials, comprising of 20 trials for 24 participants. The data represent about five hours of continuous instrumented robot control. A summary of the data available appears in Table~\ref{tab:coverage}.

\begin{table*}[t]
    
    \centering
    \begin{tabular}{lrrrr}
    \toprule
    & Left Eye & Right Eye & Egocentric Video & ZED Camera \\
    \midrule
    Total duration (h:m:s) & 5:19:26 & 5:10:45 & 5:33:44 &  4:44:45\\
    Total frames & 2299877 & 2237380 & 600728 & 512569 \\
    Nominal frequency (Hz) & 120 & 120 & 30 & 30 \\
    Frames dropped & 133301 & 195860 & 7459 & 94431 \\
    Coverage (\%) & 94.52 & 91.95 & 98.77 &  84.44  \\
    Present (\%) & 100.00 & 100.00 & 100.00 &  87.25 \\
    Coverage if present (\%) & 94.52 & 91.95 & 98.77 &  94.83 \\
    \bottomrule
    \end{tabular}\\[10pt]

    \begin{tabular}{lrrrrr}
    \toprule
     & Joystick & Robot position & Myo EMG & Myo IMU & Myo ORI \\
    \midrule
    Total duration (h:m:s) & 4:56:00 & 5:48:05 & 1:10:49 & 1:10:53 & 1:10:53 \\
    Total frames & 2131160 & 1670798 & 212465 & 212664 & 212659 \\
    Nominal frequency (Hz) & 120 & 80 & 50 & 50 & 50 \\
    Frames dropped  & 114250 & 1680 & 802368 & 802204 & 802206 \\
    Coverage (\%) & 94.91 & 99.90 & 20.94 & 20.95 & 20.95  \\
    Present (\%)  & 100.00 & 100.00 & 21.48 & 21.48 & 21.48 \\
    Coverage if present (\%) & 94.91 & 99.90 & 99.75 & 99.83 & 99.83 \\
    \bottomrule
    \end{tabular}\\[10pt]
    
    \caption{Descriptive statistics of each data stream in the data set. \emph{Total duration} and \emph{Total frames} refer to the collective amount of data of that signal over all trials and participants. \emph{Total duration} is extracted by dividing the total frames by the \emph{nominal frequency}. \emph{Frames dropped} are based on interpolating from the nominal frame rate and detecting missing data. \emph{Coverage} is computed by dividing the number of data frames by the expected number of data frames from the nominal frequency over the whole data set, \emph{Present} indicates the fraction of trials that have at least one datum of that type, and \emph{Coverage if present} is the total number of data frames divided by the expected number evaluated only if at least one datum is present in the trial.}
    \label{tab:coverage}
\end{table*}

\section{Data Streams}
\begin{figure*}
    \centering
    \includegraphics[width=.8\textwidth]{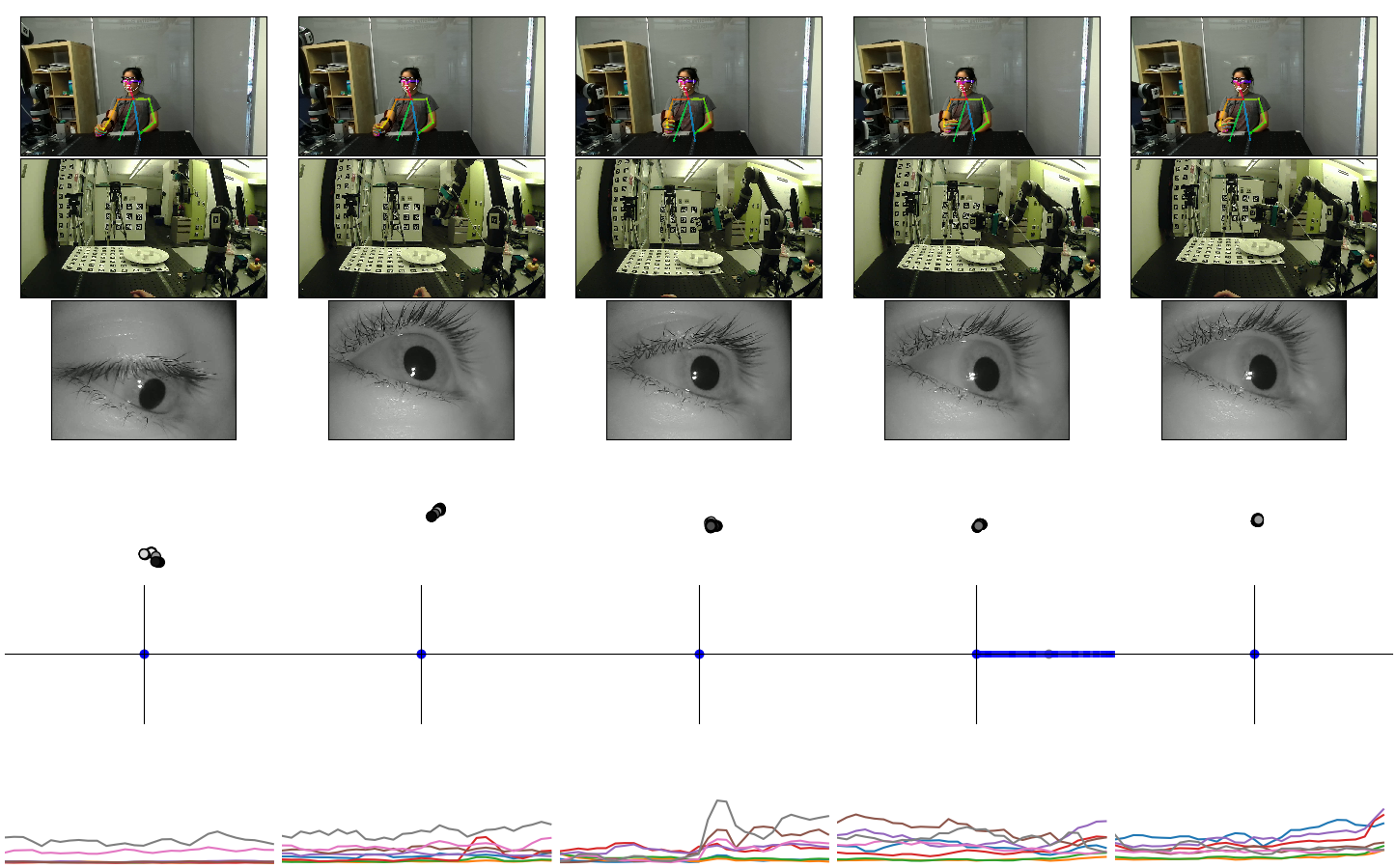}
    \caption{A visualization of several streams from the HARMONIC data set. The top row displays the ZED video with OpenPose skeletons overlaid, then the egocentric video captured from the Pupil camera, left eye video, one second of the calculated gaze dot, the trajectory of the joystick, and finally the Myo activations. For the gaze dot and the joystick, lighter colors represent more recent points in time. Each of these plots represents one second of data, sampled at 30 FPS.}
    \label{fig:staticview}
\end{figure*}

The data are organized first by participant (\texttt{p100}-\texttt{p123} reflecting the twenty-four participants). Each participant folder contains folders for three types of recordings: \texttt{calib} contains calibration passes, \texttt{check} contains intermediate gaze accuracy checks, and \texttt{run} contains data collection runs. These folders contain numbered subfolders indicating the run sequence. A visual representation of selected data streams can be seen in Fig.\ \ref{fig:staticview}.

A single trial capture (a numbered folder) has the following subfolders:
\begin{compactitem}
    \item \texttt{text\_data} contains exported \texttt{CSV} files containing the raw data. The particular raw data streams available are detailed in the following subsections. Additional to the raw data, this directory contains the body skeleton, facial, and hand keypoints generated by running OpenPose \citep{openpose_realtime, openpose_hand, openpose} on the left and right streams of the third person ZED videos. The outputs from OpenPose are compiled into face, right and left hand, and pose files for each stream of the depth camera. For full descriptions, please refer to the OpenPose documentation.
    \item \texttt{stats} contains a number of \texttt{YAML} files detailing statistical information about the trial and overall data stream, including the number of records, approximate time distances between individual records, and estimates of the times when data points may have been dropped based on the nominal data collection frame rate.
    \item \texttt{videos} contains the Pupil video files (\texttt{eye0.mp4},  \texttt{eye1.mp4} and \texttt{world.mp4}) exported as MP4 files using the H.264 video codec \citep{h264}. Additionally included are the timestamps of each frame as either numpy (\texttt{*.npy}) files, raw text (\texttt{*.txt}), or \texttt{CSV} (\texttt{*.csv}).
    \item \texttt{processed} contains a number of new formats of data extrapolated from the underlying data (e.g. a video of the egocentric recording with a dot overlaid at the gaze point). 
\end{compactitem}

\subsection{Timing and synchronization}
All data points were timestamped on collection and stored as either 32 or 64-bit floating point values in number of nanoseconds from the Unix epoch. The \texttt{CSV} files in \texttt{text\_data} provide these data in several columns.

For ease of use, two common indices are provided for all data streams. The \texttt{world\_index} field gives the  egocentric video frame number corresponding to each data point. A second common index, \texttt{world\_index\_corrected}, provides a second index into the egocentric video, with a correction for dropped video frames. The \texttt{world\_index\_corrected} value approximates a common 30Hz clock running throughout the trial. For more sophisticated data alignment, please use the provided timestamps.

\subsection{Eye Gaze}
Eye gaze videos were recorded at 120 Hz and located in the \texttt{videos} folder as \texttt{eye0.mp4} and \texttt{eye1.mp4}, encoded using the H.264 video codec \citep{h264}. Frame level timestamps are available in corresponding NumPy binary files, \texttt{eye0\_timestamps.npy} and \texttt{eye1\_timestamps.npy}. The automated pupil detection results for each eye are in the \texttt{text\_data} folder, under \texttt{pupil\_eye0.csv} and \texttt{pupil\_eye1.csv}. Field names correspond to the output of the 3D pupil detection process in Pupil Labs, as described in their documentation.

Egocentric video is available in the \texttt{videos} folder as \texttt{world.mp4} (encoded using the H.264 codec \citep{h264}), with frame level timestamps located in \texttt{world\_timestamps.npy}. Calculated gaze position within the corresponding video frame is given in \texttt{text\_data/gaze\_positions.csv}. See the pupil labs documentation for a full description of fields. The fields \texttt{norm\_pos\_x} and \texttt{norm\_pos\_y} correspond to the $(x,y)$ pixel in coordinates normalized to the egocentric video frame size, with the origin point in the top left.

Data used to calibrate between pupil data and gaze point are stored in the text files \texttt{pupil\_cal\_eye0.csv}, \texttt{pupil\_cal\_eye1.csv}, and \texttt{world\_cal\_positions.csv}. These data are the same between runs of the same participant and is provided as a convenience to recalculate a calibration if desired. Details of the current calibration method can be found in the Pupil Labs software documentation.

\subsection{Third Person Video}
ZED videos were recorded using the Stereolabs ZED software, version 1.1.0. Data were initially stored as a Stereolabs SVO file, including separate left and right videos and a common timestamp. Videos were extracted to the \texttt{videos} directory as \texttt{zed\_left.mp4} and \texttt{zed\_right.mp4} encoded using H.264 \citep{h264}. The timestamps were rescaled to the Unix epoch and stored as an integer number of nanoseconds from the epoch in \texttt{zed\_ts.txt}, as well as floating-point NumPy format in \texttt{zed\_timestamps.npy}. The \texttt{zed\_corrs.csv} stores the correlations to a common index, as previously explained.

\subsection{Additional sensor data}
The following data streams are available in the \texttt{text\_data} directory, having been extracted or calculated from the original binary. 

\begin{compactitem}
    \item \texttt{control\_mode.txt} contains one character referring to that trial's assistance condition. Zero represents direct teleoperation and 3 represents robot control.
    \item \texttt{morsel.yaml} is a \texttt{YAML} file with the transforms for each detected morsel positions in the robot base frame.
    \item \texttt{ada\_joy.csv} stores raw joystick input provided by the user. Joystick input is only provided when changed from the previous message leading to inconsistent timing in the raw data. To rectify this, joystick data have been resampled to a common 120 Hz frequency and missing data filled by the previous value. Duplicate data are noted by unchanged headers.
    \item \texttt{input\_info.csv} contains the user input to the robot. The \texttt{robot\_mode} field denotes which control mode the robot is in (x/y, z/yaw, or pitch/roll), and the rest of the fields denote the applied twist corresponding to the user's joystick input.
    \item \texttt{assistance\_info.csv} contains the outcome of the shared autonomy algorithm. It stores the current probability inferred for each goal and the resultant twist applied to the robot at that timestep.
    \item \texttt{joint\_states.csv} contains the information for each joint of the robot.
    \item \texttt{robot\_position.csv} contains the cartesian position of each of the robot links, as calculated from the forward kinematics using the data from \texttt{joint\_states.csv}.
    \item \texttt{myo\_emg.csv} contains EMG output of the Myo.
    \item \texttt{myo\_imu.csv} contains IMU output of the Myo.
    \item \texttt{myo\_ori.csv} contains orientation data received from the Myo sensor.
\end{compactitem}

\section{Known Issues}
\subsection{Missing Data}
Due to computational load, certain data streams may have periodic dropouts. The \texttt{stats} directory contains some info on when and how often these occur, and overall statistics are given in Table~\ref{tab:coverage}. The missing data are particularly exacerbated for the Myo signal due to the data recording software failing to start. Finally, due to permissions restrictions, unedited ZED video capture is available for 10 participants, de-identified video (video with faces blurred) is available for 13 participants, and video for 1 participant is unavailable for release. Within the released participants, some initialization failure means that videos of certain trials are occasionally missing.

\section{Accessing the Data}
The data will be hosted on the HARP Lab website: \url{http://harp.ri.cmu.edu/harmonic}. Several files are provided for download: \texttt{harmonic\_data.tar.gz}, a compilation of all of the data, ($\sim$ 68 Gb), \texttt{harmonic\_minimal.tar.gz}, consisting of the \texttt{text\_data}, \texttt{videos}, and \texttt{stats} directories,  ($\sim$ 15 Gb), \texttt{harmonic\_text.tar.gz}, consisting of the \texttt{text\_data} directory, ($\sim$ 4 Gb), and finally \texttt{harmonic\_sample.tar.gz}, consisting of all of the data for a single participant, ($\sim$ 303 Mb). The data sets will be versioned using semantic versioning, and that page will maintain a log of all changes that may be made to the data set after release. Furthermore, our GitHub contains a repository for basic processing tools located here: \url{https://github.com/HARPLab/harmonic_cpp}. Finally, for the original robot control code, follow this link to a fork of the publicly available implementation of the shared autonomy code we used: \url{https://github.com/HARPLab/ada_meal_scenario} \citep{javdani15}. Our robot control code is on the branch: ``adjustable". 

\section{Conclusion}

We presented a data set of humans who performed  a food acquisition task by controlling a robot manipulator. During this task, a variety of types of participant data were collected, including eye gaze information, electrymyography of the controlling arm, stereo video, and robot controller information. This data set enables research into human-robot collaboration and multimodal human behavior analysis.

\section{Acknowledgements}
This work was supported by the National Science Foundation (IIS-1755823) and the Paralyzed Veterans of America. The first author is supported by a National Science Foundation Graduate Research Fellowship (DGE 1745016).

\section{Conflicts of Interest}
Siddhartha Srinivasa is a Multimedia Editor at the International Journal of Robotics Research (IJRR). The authors declare no other conflicts of interest.
\bibliographystyle{SageH}
\bibliography{refs/longnames,refs/master,refs/bib}

\begin{thebibliography}{48}
\providecommand{\natexlab}[1]{#1}
\providecommand{\url}[1]{\texttt{#1}}
\providecommand{\urlprefix}{URL }
\expandafter\ifx\csname urlstyle\endcsname\relax
  \providecommand{\doi}[1]{DOI:\discretionary{}{}{}#1}\else
  \providecommand{\doi}{DOI:\discretionary{}{}{}\begingroup
  \urlstyle{rm}\Url}\fi

\bibitem[{Admoni et~al.(2014)Admoni, Datsikas and Scassellati}]{admoni14cogsci}
Admoni H, Datsikas C and Scassellati B (2014) Speech and gaze conflicts in
  collaborative human-robot interactions.
\newblock In: \emph{Annual Conference of the Cognitive Science Society
  ({CogSci})}. pp. 104--109.

\bibitem[{Aronson and Admoni(2018)}]{aronson18rss_fja}
Aronson RM and Admoni H (2018) Gaze for error detection during human-robot
  shared manipulation.
\newblock In: \emph{RSS Workshop: Towards a Framework for Joint Action}.

\bibitem[{Aronson et~al.(2018)Aronson, Santini, Kubler, Kasneci, Srinivasa and
  Admoni}]{aronson18}
Aronson RM, Santini T, Kubler TC, Kasneci E, Srinivasa SS and Admoni H (2018)
  Eye-hand behavior in human-robot shared manipulation.
\newblock In: \emph{{ACM/IEEE} International Conference on Human-Robot
  Interaction ({HRI})}.

\bibitem[{Artemiadis and Kyriakopoulos(2010)}]{Artemiadis10}
Artemiadis PK and Kyriakopoulos KJ (2010) Emg-based control of a robot arm
  using low-dimensional embeddings.
\newblock \emph{IEEE Transactions on Robotics} 26(2): 393--398.
\newblock \doi{10.1109/TRO.2009.2039378}.

\bibitem[{Azagra et~al.(2016)Azagra, Mollard, Golemo, Murillo, Lopes and
  Civera}]{azgara16}
Azagra P, Mollard Y, Golemo F, Murillo AC, Lopes M and Civera J (2016) {A
  Multimodal Human-Robot Interaction Dataset}.
\newblock {NIPS 2016, workshop Future of Interactive Learning Machines}.
\newblock \urlprefix\url{https://hal.inria.fr/hal-01402479}.
\newblock Poster.

\bibitem[{Bastianelli et~al.(2014)Bastianelli, Castellucci, Croce, Iocchi,
  Basili and Nardi}]{bastianelli14}
Bastianelli E, Castellucci G, Croce D, Iocchi L, Basili R and Nardi D (2014)
  Huric: a human robot interaction corpus.
\newblock In: Chair) NCC, Choukri K, Declerck T, Loftsson H, Maegaard B,
  Mariani J, Moreno A, Odijk J and Piperidis S (eds.) \emph{Proceedings of the
  Ninth International Conference on Language Resources and Evaluation
  (LREC'14)}. Reykjavik, Iceland: European Language Resources Association
  (ELRA).
\newblock ISBN 978-2-9517408-8-4.

\bibitem[{Beatty(1982)}]{beatty82}
Beatty J (1982) Task-evoked pupillary responses, processing load, and the
  structure of processing resources.
\newblock \emph{Psychological Bulletin} 91(2): 276.

\bibitem[{Bednarik et~al.(2018)Bednarik, Bartczak, Vrzakova, Koskinen, Elomaa,
  Huotarinen, de~G\'{o}mez~P{\'e}rez and von und~zu Fraunberg}]{Bednarik2018}
Bednarik R, Bartczak P, Vrzakova H, Koskinen J, Elomaa AP, Huotarinen A,
  de~G\'{o}mez~P{\'e}rez DG and von und~zu Fraunberg M (2018) Pupil size as an
  indicator of visual-motor workload and expertise in microsurgical training
  tasks.
\newblock In: \emph{Proceedings of the 2018 ACM Symposium on Eye Tracking
  Research \& Applications}, ETRA '18. New York, NY, USA: ACM.
\newblock ISBN 978-1-4503-5706-7, pp. 60:1--60:5.
\newblock \doi{10.1145/3204493.3204577}.
\newblock \urlprefix\url{http://doi.acm.org/10.1145/3204493.3204577}.

\bibitem[{Bellemare et~al.(2012)Bellemare, Naddaf, Veness and Bowling}]{atari}
Bellemare MG, Naddaf Y, Veness J and Bowling M (2012) The arcade learning
  environment: An evaluation platform for general agents.
\newblock \emph{CoRR} abs/1207.4708.
\newblock \urlprefix\url{http://arxiv.org/abs/1207.4708}.

\bibitem[{Ben-Youssef et~al.(2017)Ben-Youssef, Clavel, Essid, Bilac, Chamoux
  and Lim}]{benyoussef17}
Ben-Youssef A, Clavel C, Essid S, Bilac M, Chamoux M and Lim A (2017) Ue-hri: A
  new dataset for the study of user engagement in spontaneous human-robot
  interactions.
\newblock In: \emph{Proceedings of the 19th ACM International Conference on
  Multimodal Interaction}, ICMI 2017. New York, NY, USA: ACM.
\newblock ISBN 978-1-4503-5543-8, pp. 464--472.
\newblock \doi{10.1145/3136755.3136814}.
\newblock \urlprefix\url{http://doi.acm.org/10.1145/3136755.3136814}.

\bibitem[{Boucher et~al.(2012)Boucher, Pattacini, Lelong, Bailly, Elisei,
  Fagel, Dominey and Ventre-Dominey}]{boucher12}
Boucher JD, Pattacini U, Lelong A, Bailly G, Elisei F, Fagel S, Dominey PF and
  Ventre-Dominey J (2012) {I Reach Faster When I See You Look: Gaze Effects in
  Human-Human and Human-Robot Face-to-Face Cooperation.}
\newblock \emph{Frontiers in neurorobotics} 6(May): 1--11.
\newblock \doi{10.3389/fnbot.2012.00003}.

\bibitem[{Braunagel et~al.(2015)Braunagel, Kasneci, Stolzmann and
  Rosenstiel}]{Braunagel15}
Braunagel C, Kasneci E, Stolzmann W and Rosenstiel W (2015) Driver-activity
  recognition in the context of conditionally autonomous driving.
\newblock In: \emph{2015 IEEE 18th International Conference on Intelligent
  Transportation Systems}. pp. 1652--1657.
\newblock \doi{10.1109/ITSC.2015.268}.

\bibitem[{Cao et~al.(2017)Cao, Simon, Wei and Sheikh}]{openpose_realtime}
Cao Z, Simon T, Wei SE and Sheikh Y (2017) Realtime multi-person 2d pose
  estimation using part affinity fields.
\newblock In: \emph{CVPR}.

\bibitem[{Damen et~al.(2018)Damen, Doughty, Farinella, Fidler, Furnari,
  Kazakos, Moltisanti, Munro, Perrett, Price and Wray}]{epic}
Damen D, Doughty H, Farinella GM, Fidler S, Furnari A, Kazakos E, Moltisanti D,
  Munro J, Perrett T, Price W and Wray M (2018) Scaling egocentric vision: The
  {EPIC-KITCHENS} dataset.
\newblock \emph{CoRR} abs/1804.02748.
\newblock \urlprefix\url{http://arxiv.org/abs/1804.02748}.

\bibitem[{DelPreto et~al.(2018)DelPreto, Salazar-Gomez, Gil, Hasani, Guenther
  and Rus}]{DelPreto18}
DelPreto J, Salazar-Gomez AF, Gil S, Hasani RM, Guenther FH and Rus D (2018)
  Plug-and-play supervisory control using muscle and brain signals for
  real-time gesture and error detection.
\newblock In: \emph{Robotics: Science and Systems}.

\bibitem[{Fathi et~al.(2012)Fathi, Li and Rehg}]{gtea}
Fathi A, Li Y and Rehg JM (2012) Learning to recognize daily actions using
  gaze.
\newblock In: Fitzgibbon A, Lazebnik S, Perona P, Sato Y and Schmid C (eds.)
  \emph{Computer Vision -- ECCV 2012}. Berlin, Heidelberg: Springer Berlin
  Heidelberg.
\newblock ISBN 978-3-642-33718-5, pp. 314--327.

\bibitem[{Hayhoe and Ballard(2005)}]{hayhoe05}
Hayhoe M and Ballard D (2005) Eye movements in natural behavior.
\newblock \emph{Trends in Cognitive Sciences} 9(4): 188--194.
\newblock \doi{10.1016/j.tics.2005.02.009}.

\bibitem[{Huang and Mutlu(2016)}]{huang16}
Huang CM and Mutlu B (2016) Anticipatory robot control for efficient
  human-robot collaboration.
\newblock In: \emph{{ACM/IEEE} International Conference on Human-Robot
  Interaction ({HRI})}. pp. 83--90.

\bibitem[{Javdani et~al.(2018)Javdani, Admoni, Pellegrinelli, Srinivasa and
  Bagnell}]{javdani17ijrr}
Javdani S, Admoni H, Pellegrinelli S, Srinivasa SS and Bagnell JA (2018) Shared
  autonomy via hindsight optimization for teleoperation and teaming.
\newblock \emph{IJRR} .

\bibitem[{Javdani et~al.(2015)Javdani, Srinivasa and Bagnell}]{javdani15}
Javdani S, Srinivasa SS and Bagnell JA (2015) Shared autonomy via hindsight
  optimization.
\newblock In: \emph{Robotics: Science and Systems ({RSS})}.

\bibitem[{Jayagopi et~al.(2013)Jayagopi, Sheiki, Klotz, Wienke, Odobez, Wrede,
  Khalidov, Nyugen, Wrede and Gatica-Perez}]{jayagopi13}
Jayagopi DB, Sheiki S, Klotz D, Wienke J, Odobez J, Wrede S, Khalidov V, Nyugen
  L, Wrede B and Gatica-Perez D (2013) The vernissage corpus: A conversational
  human-robot-interaction dataset.
\newblock In: \emph{2013 8th ACM/IEEE International Conference on Human-Robot
  Interaction (HRI)}. pp. 149--150.
\newblock \doi{10.1109/HRI.2013.6483545}.

\bibitem[{Johansson et~al.(2001)Johansson, Westling, {B{\"{a}} Ckstr{\"{o}}}
  and Flanagan}]{Johansson2001}
Johansson RS, Westling GR, {B{\"{a}} Ckstr{\"{o}}} A and Flanagan JR (2001)
  {Eye--Hand Coordination in Object Manipulation}.
\newblock \emph{The Journal of Neuroscience} 21(17): 6917--6932.

\bibitem[{Kaelbling et~al.(1998)Kaelbling, Littman and
  Cassandra}]{KAELBLING199899}
Kaelbling LP, Littman ML and Cassandra AR (1998) Planning and acting in
  partially observable stochastic domains.
\newblock \emph{Artificial Intelligence} 101(1): 99 -- 134.
\newblock \doi{https://doi.org/10.1016/S0004-3702(98)00023-X}.
\newblock
  \urlprefix\url{http://www.sciencedirect.com/science/article/pii/S000437029800023X}.

\bibitem[{Kassner et~al.(2014)Kassner, Patera and Bulling}]{PupilLabsAcademic}
Kassner M, Patera W and Bulling A (2014) Pupil: An open source platform for
  pervasive eye tracking and mobile gaze-based interaction.
\newblock In: \emph{Adjunct Proceedings of the 2014 ACM International Joint
  Conference on Pervasive and Ubiquitous Computing}, UbiComp '14 Adjunct. New
  York, NY, USA: ACM.
\newblock ISBN 978-1-4503-3047-3, pp. 1151--1160.
\newblock \doi{10.1145/2638728.2641695}.
\newblock \urlprefix\url{http://doi.acm.org/10.1145/2638728.2641695}.

\bibitem[{Krejtz et~al.(2018)Krejtz, Duchowski, Niedzielska, Biele and
  Krejtz}]{Krejtz2018}
Krejtz K, Duchowski AT, Niedzielska A, Biele C and Krejtz I (2018) Eye tracking
  cognitive load using pupil diameter and microsaccades with fixed gaze.
\newblock \emph{PLOS ONE} 13(9): 1--23.
\newblock \doi{10.1371/journal.pone.0203629}.
\newblock \urlprefix\url{https://doi.org/10.1371/journal.pone.0203629}.

\bibitem[{Land and Hayhoe(2001{\natexlab{a}})}]{land01}
Land MF and Hayhoe M (2001{\natexlab{a}}) {In what ways do eye movements
  contribute to everyday activities?}
\newblock \emph{Vision Research} 41(25-26): 3559--65.

\bibitem[{Land and Hayhoe(2001{\natexlab{b}})}]{Land2001}
Land MF and Hayhoe M (2001{\natexlab{b}}) {In what ways do eye movements
  contribute to everyday activities?}
\newblock \emph{Vision Research} 41(25): 3559--3565.

\bibitem[{Littman et~al.(1995)Littman, Cassandra and
  Kaelbling}]{LITTMAN1995362}
Littman ML, Cassandra AR and Kaelbling LP (1995) Learning policies for
  partially observable environments: Scaling up.
\newblock In: Prieditis A and Russell S (eds.) \emph{Machine Learning
  Proceedings 1995}. San Francisco (CA): Morgan Kaufmann.
\newblock ISBN 978-1-55860-377-6, pp. 362 -- 370.
\newblock \doi{https://doi.org/10.1016/B978-1-55860-377-6.50052-9}.
\newblock
  \urlprefix\url{http://www.sciencedirect.com/science/article/pii/B9781558603776500529}.

\bibitem[{Mainprice et~al.(2015)Mainprice, Hayne and Berenson}]{Mainprice2015}
Mainprice J, Hayne R and Berenson D (2015) Predicting human reaching motion in
  collaborative tasks using inverse optimal control and iterative re-planning.
\newblock In: \emph{2015 IEEE International Conference on Robotics and
  Automation (ICRA)}. pp. 885--892.
\newblock \doi{10.1109/ICRA.2015.7139282}.

\bibitem[{Pirsiavash and Ramanan(2012)}]{adl}
Pirsiavash H and Ramanan D (2012) Detecting activities of daily living in
  first-person camera views.
\newblock In: \emph{2012 IEEE Conference on Computer Vision and Pattern
  Recognition}. pp. 2847--2854.
\newblock \doi{10.1109/CVPR.2012.6248010}.

\bibitem[{{Pupil Labs, Inc.}(2017)}]{pupillabswebsite}
{Pupil Labs, Inc} (2017) Pupil labs - pupil.
\newblock \url{https://pupil-labs.com/pupil/}.

\bibitem[{Reddy et~al.(2018{\natexlab{a}})Reddy, Dragan and
  Levine}]{reddyARXIV18}
Reddy S, Dragan AD and Levine S (2018{\natexlab{a}}) Where do you think you're
  going?: Inferring beliefs about dynamics from behavior.
\newblock \emph{CoRR} abs/1805.08010.
\newblock \urlprefix\url{http://arxiv.org/abs/1805.08010}.

\bibitem[{Reddy et~al.(2018{\natexlab{b}})Reddy, Levine and
  Dragan}]{reddyRSS18}
Reddy S, Levine S and Dragan AD (2018{\natexlab{b}}) Shared autonomy via deep
  reinforcement learning.
\newblock \emph{CoRR} abs/1802.01744.
\newblock \urlprefix\url{http://arxiv.org/abs/1802.01744}.

\bibitem[{Richardson(2010)}]{h264}
Richardson IE (2010) \emph{The H.264 Advanced Video Compression Standard}.
\newblock 2nd edition. Wiley Publishing.
\newblock ISBN 0470516925.

\bibitem[{Sheikhi and Odobez(2012)}]{sheikhi12}
Sheikhi S and Odobez JM (2012) Recognizing the visual focus of attention for
  human robot interaction.
\newblock In: \emph{Proceedings of the Third International Conference on Human
  Behavior Understanding}, HBU'12. Berlin, Heidelberg: Springer-Verlag.
\newblock ISBN 978-3-642-34013-0, pp. 99--112.
\newblock \doi{10.1007/978-3-642-34014-7_9}.
\newblock \urlprefix\url{http://dx.doi.org/10.1007/978-3-642-34014-7_9}.

\bibitem[{Shu et~al.(2016)Shu, Ryoo and Zhu}]{shu16}
Shu T, Ryoo MS and Zhu SC (2016) Learning social affordance for human-robot
  interaction.
\newblock In: \emph{International Joint Conference on Artificial Intelligence
  (IJCAI)}.

\bibitem[{Sigurdsson et~al.(2018{\natexlab{a}})Sigurdsson, Gupta, Schmid,
  Farhadi and Alahari}]{char}
Sigurdsson GA, Gupta A, Schmid C, Farhadi A and Alahari K (2018{\natexlab{a}})
  Charades-ego: {A} large-scale dataset of paired third and first person
  videos.
\newblock \emph{CoRR} abs/1804.09626.
\newblock \urlprefix\url{http://arxiv.org/abs/1804.09626}.

\bibitem[{Sigurdsson et~al.(2018{\natexlab{b}})Sigurdsson, Gupta, Schmid,
  Farhadi and Alahari}]{char-ego}
Sigurdsson GA, Gupta A, Schmid C, Farhadi A and Alahari K (2018{\natexlab{b}})
  Charades-ego: {A} large-scale dataset of paired third and first person
  videos.
\newblock \emph{CoRR} abs/1804.09626.
\newblock \urlprefix\url{http://arxiv.org/abs/1804.09626}.

\bibitem[{Simon et~al.(2017)Simon, Joo, Matthews and Sheikh}]{openpose_hand}
Simon T, Joo H, Matthews I and Sheikh Y (2017) Hand keypoint detection in
  single images using multiview bootstrapping.
\newblock In: \emph{CVPR}.

\bibitem[{Sondik(1978)}]{sondick}
Sondik EJ (1978) The optimal control of partially observable markov processes
  over the infinite horizon: Discounted costs.
\newblock \emph{Operations Research} 26(2): 282--304.
\newblock \urlprefix\url{http://www.jstor.org/stable/169635}.

\bibitem[{Stefanov and Beskow(2016)}]{stefanov16}
Stefanov K and Beskow J (2016) A multi-party multi-modal dataset for focus of
  visual attention in human-human and human-robot interaction.
\newblock In: \emph{LREC}.

\bibitem[{{Stereolabs Inc.}(2018)}]{stereolabswebsite}
{Stereolabs Inc} (2018) {Stereolabs}.
\newblock \url{https://www.stereolabs.com/}.

\bibitem[{{Thalmic Labs, Inc.}(2018)}]{myowebsite}
{Thalmic Labs, Inc} (2018) {Myo Gesture Control Armband}.
\newblock \url{https://www.myo.com/}.

\bibitem[{Tobii(2017)}]{tobiiwebsite}
Tobii I (2017) Tobii pro website.

\bibitem[{Wei et~al.(2016)Wei, Ramakrishna, Kanade and Sheikh}]{openpose}
Wei SE, Ramakrishna V, Kanade T and Sheikh Y (2016) Convolutional pose
  machines.
\newblock In: \emph{CVPR}.

\bibitem[{Xu et~al.(2013)Xu, Zhang and Yu}]{xu13}
Xu TL, Zhang H and Yu C (2013) Cooperative gazing behaviors in human
  multi-robot interaction.
\newblock \emph{Interaction Studies} 14: 390--418.

\bibitem[{Zhang et~al.(2018)Zhang, Liu, Zhang, Whritner, Muller, Hayhoe and
  Ballard}]{zhangECCV18}
Zhang R, Liu Z, Zhang L, Whritner JA, Muller KS, Hayhoe MM and Ballard DH
  (2018) {AGIL:} learning attention from human for visuomotor tasks.
\newblock In: \emph{Computer Vision - {ECCV} 2018 - 15th European Conference,
  Munich, Germany, September 8-14, 2018, Proceedings, Part {XI}}. pp. 692--707.
\newblock \doi{10.1007/978-3-030-01252-6\_41}.
\newblock \urlprefix\url{https://doi.org/10.1007/978-3-030-01252-6\_41}.

\bibitem[{Ziebart et~al.(2008)Ziebart, Maas, Bagnell and
  Dey}]{ziebart2008maximum}
Ziebart BD, Maas A, Bagnell JA and Dey AK (2008) Maximum entropy inverse
  reinforcement learning.
\newblock In: \emph{Proc. AAAI}. pp. 1433--1438.

\end{thebibliography}

\end{document}